\documentclass[conference]{IEEEtran}
\usepackage{times}
\usepackage{float}

\usepackage[numbers]{natbib}
\usepackage{multicol}
\usepackage{caption}
\usepackage{graphicx}
\usepackage{color}
\usepackage{amsmath}
\usepackage[bookmarks=true]{hyperref}

\begin{document}

\title{Deep Image Homography Estimation}


\author{\authorblockN{Daniel DeTone}
\authorblockA{Magic Leap, Inc.\\
Mountain View, CA\\
ddetone@magicleap.com}
\and
\authorblockN{Tomasz Malisiewicz}
\authorblockA{Magic Leap, Inc.\\
Mountain View, CA\\
tmalisiewicz@magicleap.com}
\and
\authorblockN{Andrew Rabinovich}
\authorblockA{Magic Leap, Inc.\\
Mountain View, CA\\
arabinovich@magicleap.com}}

\maketitle

\begin{abstract}
We present a deep convolutional neural network for estimating the relative
homography between a pair of
images. Our feed-forward network has 10 layers, takes two stacked grayscale images as
input, and produces an 8 degree of freedom homography which can be used to map the pixels from the first image to the second. We present two convolutional neural network architectures for HomographyNet: a regression network which directly estimates the
real-valued homography parameters, and a classification network which produces 
a distribution over quantized homographies. We use a 4-point homography parameterization
which maps the four corners from one image into
the second image.  Our networks are trained in an end-to-end 
fashion using warped MS-COCO images. Our approach works without the need for separate
local feature detection and transformation estimation stages. Our deep models are
compared to a traditional homography estimator based on ORB features
and we highlight the scenarios where HomographyNet outperforms
the traditional technique. We also describe a variety of applications powered by deep
homography estimation, thus showcasing the flexibility of a deep
learning approach.
\end{abstract}

\IEEEpeerreviewmaketitle

\section{Introduction}

Sparse 2D feature points are the basis of most modern Structure from
Motion and SLAM techniques~\cite{Hartley04}. These sparse 2D features are typically
known as corners, and in all geometric computer vision tasks one must
balance the errors in corner detection methods with geometric
estimation errors. Even the simplest geometric methods, like
estimating the homography between two images, rely on the error-prone
corner-detection method.

Estimating a 2D homography (or projective transformation) from a pair
of images is a fundamental task in computer vision. The homography is
an essential part of monocular SLAM systems in scenarios such as:

\begin{itemize}
  \item Rotation only movements
  \item Planar scenes
  \item Scenes in which objects are very far from the viewer
\end{itemize}

It is well-known that the transformation relating two images
undergoing a rotation about the camera center is a homography, and it
is not surprising that homographies are essential for creating panoramas~\cite{brown2007automatic}. To deal with planar and mostly-planar scenes, the popular SLAM algorithm ORB-SLAM~\cite{orbslam} uses a combination of homography estimation and fundamental matrix estimation. Augmented Reality applications based on planar structures and homographies have been well-studied~\cite{Simon00}. Camera calibration techniques using planar structures~\cite{zhang2000flexible} also rely on homographies.

The traditional homography estimation pipeline is composed of two
stages: corner estimation and robust homography estimation. Robustness
is introduced into the corner detection stage by returning a large and
over-complete set of points, while robustness into the homography
estimation step shows up as heavy use of RANSAC or robustification of the squared loss function. Since corners are not as reliable as man-made linear
structures, the research community has put considerable effort into adding line features~\cite{SRD06} and more complicated geometries~\cite{Gee08}
into the feature detection step. What we really want is a single
robust algorithm that, given a pair of images, simply returns the
homography relating the pair. \emph{Instead of manually engineering
corner-ish features, line-ish features, etc, is it possible for the
algorithm to learn its own set of primitives?} We want to go even further, and add the transformation estimation step as the last part of a
deep learning pipeline, thus giving us the ability to learn the entire homography estimation pipeline in an end-to-end fashion.

\begin{figure*}
\centering
\includegraphics[width=\textwidth]{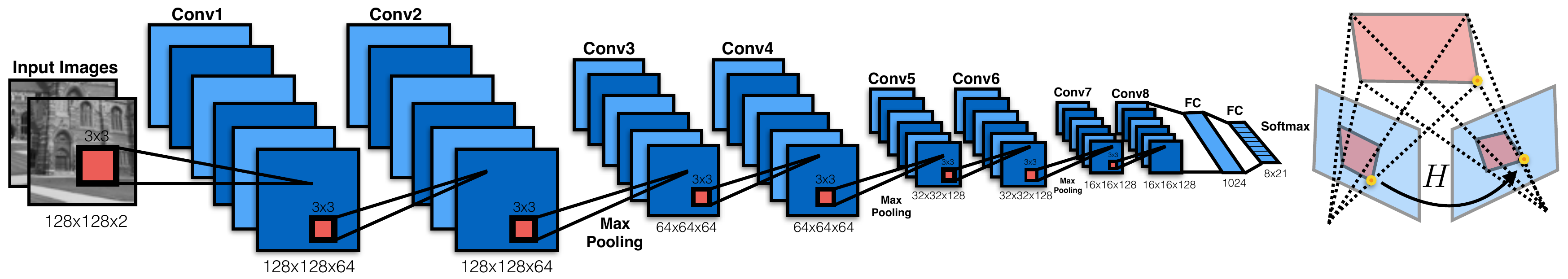}
\caption{{\bf Deep Image Homography Estimation.} HomographyNet is a Deep Convolutional Neural Network which directly produces the Homography
  relating two images. Our method does net require separate
  corner detection and homography estimation steps and all parameters
  are trained in an end-to-end fashion using a large dataset of
  labeled images. \label{fig:teaser}} 
\end{figure*}

Recent research in “dense” or “direct” featureless SLAM algorithms
such as LSD-SLAM~\cite{engel14eccv} indicates promise in using a full
image for geometric computer vision tasks. Concurrently, deep
convolutional networks are setting state-of-the-art benchmarks in
semantic tasks such as image classification, semantic segmentation and
human pose estimation. Additionally, recent works such as
FlowNet~\cite{FlowNet15}, Deep Semantic Matching~\cite{Bai16} and
Eigen \emph{et al.}'s Multi-Scale Deep Network~\cite{Eigen14} present promising results for dense
geometric computer vision tasks like optical flow and depth
estimation. Even robotic tasks like visual odometry are being tackled with convolutional neural networks~\cite{Constante2016}.

In this paper, we show that the entire homography estimation problem can
be solved by a deep convolutional neural network (See Figure \ref{fig:teaser}). Our contributions are as follows: we present a new
VGG-style~\cite{Simonyan14} network for the homography estimation
task. We show how to use the 4-point parameterization~\cite{Baker06} to
get a well-behaved deep estimation problem. Because deep networks require a lot of data to be trained from scratch, we share our recipe for creating a seemingly infinite dataset of $(I_A, I_B, H^{AB})$ training triplets from an existing dataset of real images like the
MS-COCO dataset. We present an additional formulation of the
homography estimation problem as classification, which produces a
distribution over homographies and can be used to determine the
confidence of an estimated homography.

\section{The 4-point Homography Parameterization}
 
The simplest way to parameterize a homography is with a 3x3 matrix and a fixed scale. The homography maps $[u,v]$, the pixels in the left image, to $[u',v']$, the pixels in the right image, and is defined up to scale (see Equation~\ref{eqn:homography}). 

\begin{equation}
\begin{pmatrix} u' \\ v' \\ 1 \end{pmatrix} \sim \begin{pmatrix} H_{11} & H_{12} & H_{13} \\ H_{21} & H_{22} & H_{23} \\ H_{31} & H_{32} & H_{33} \end{pmatrix} \begin{pmatrix} u \\ v \\ 1 \end{pmatrix}
\label{eqn:homography}
\end{equation}

However, if we unroll the 8 (or 9) parameters of the
homography into a single vector, we'll quickly realize that we are
mixing both rotational and translational terms. For example, the submatrix $[H_{11}\ H_{12};\ H_{21}\ H_{22}]$, represents the rotational terms in the homography, while the vector $[H_{13}\ H_{23}]$ is the translational offset. Balancing the
rotational and translational terms as part of an optimization problem is difficult.

We found that an alternate parameterization, one based on a single kind
of “location” variable, namely the corner location, is more suitable for our deep homography estimation task. The 4-point
parameterization has been used in traditional homography estimation
methods~\cite{Baker06}, and we use it in our modern deep manifestation
of the homography estimation problem (See Figure~\ref{fig:4point}). Letting $\Delta u_1 = u_1' - u_1$ be the u-offset for the first corner, the 4-point parameterization represents a homography as follows:

\begin{equation}
H_{4point} = \begin{pmatrix} \Delta u_{1} & \Delta v_{1} \\  \Delta u_{2} & \Delta v_{2}   \\  \Delta u_{3} & \Delta v_{3} \\  \Delta u_{4} & \Delta v_{4}  \end{pmatrix} 
\end{equation}

Equivalently to the matrix formulation of the homography, the 4-point parameterization uses eight numbers. Once the displacement of the four corners is known, one can easily convert $H_{4point}$ to $H_{matrix}$. This can be accomplished in a number of ways,
for example one can use the normalized Direct Linear Transform (DLT)
algorithm \cite{Hartley04}, or the function {\verb getPerspectiveTransform() in OpenCV.

\section{Data Generation for Homography Estimation}
\label{sec:data}
Training deep convolutional networks from scratch requires a large
amount of data. To meet this requirement, we generate a nearly
unlimited number of labeled training examples by applying random
projective transformations to a large dataset of natural images
\footnote{In our experiments, we used cropped MS-COCO~\cite{coco14}
images, although any large-enough dataset could be used for training}.
The process is illustrated in Figure \ref{fig:data} and described below.

\begin{figure}[b]
\centering
\includegraphics[width=.45\textwidth]{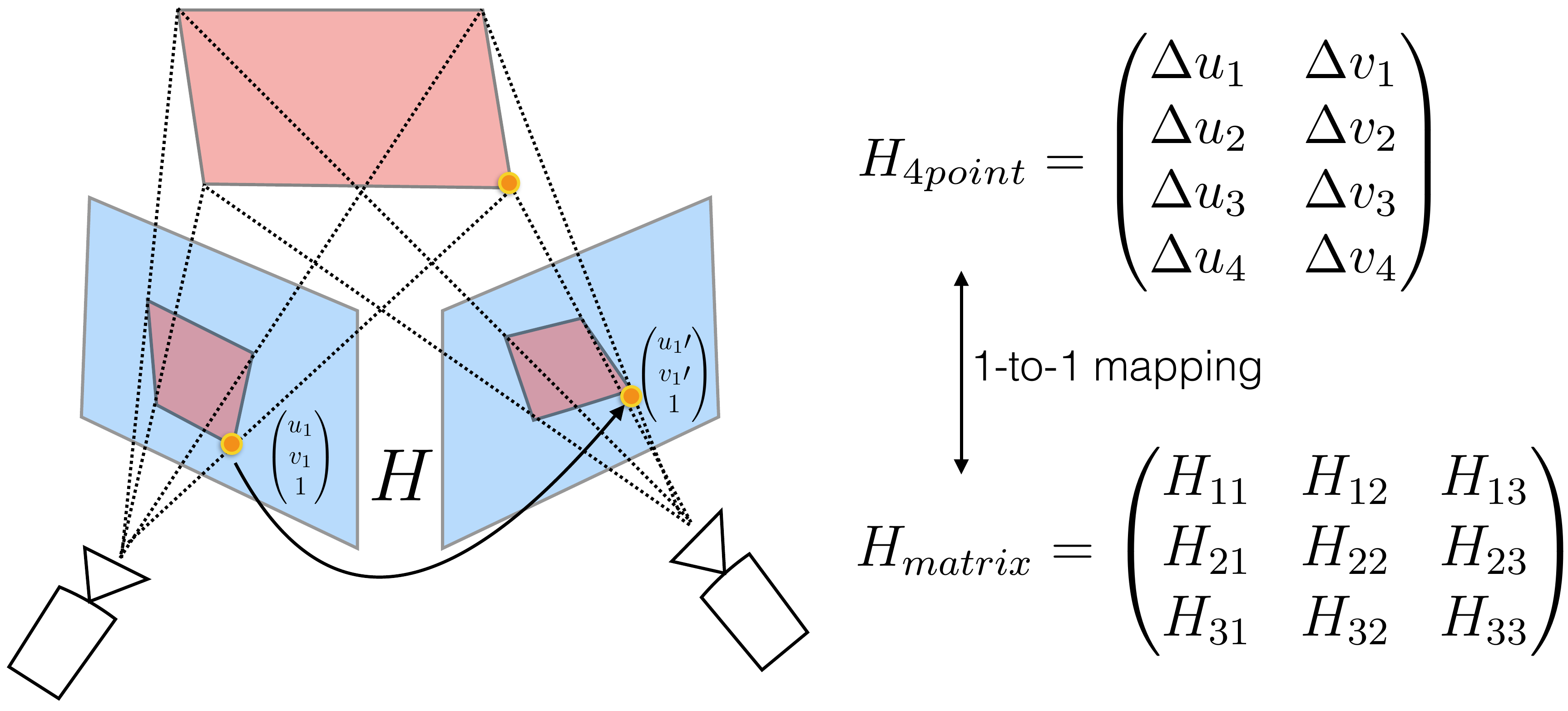}
\caption{{\bf 4-point parameterization.} We use the 4-point parameterization of the homography. There exists a 1-to-1 mapping between the 8-dof "corner offset" matrix and the representation of the homography as a 3x3 matrix. \label{fig:4point}}
\end{figure}

\begin{figure}[ht]
\centering
\includegraphics[width=.45\textwidth]{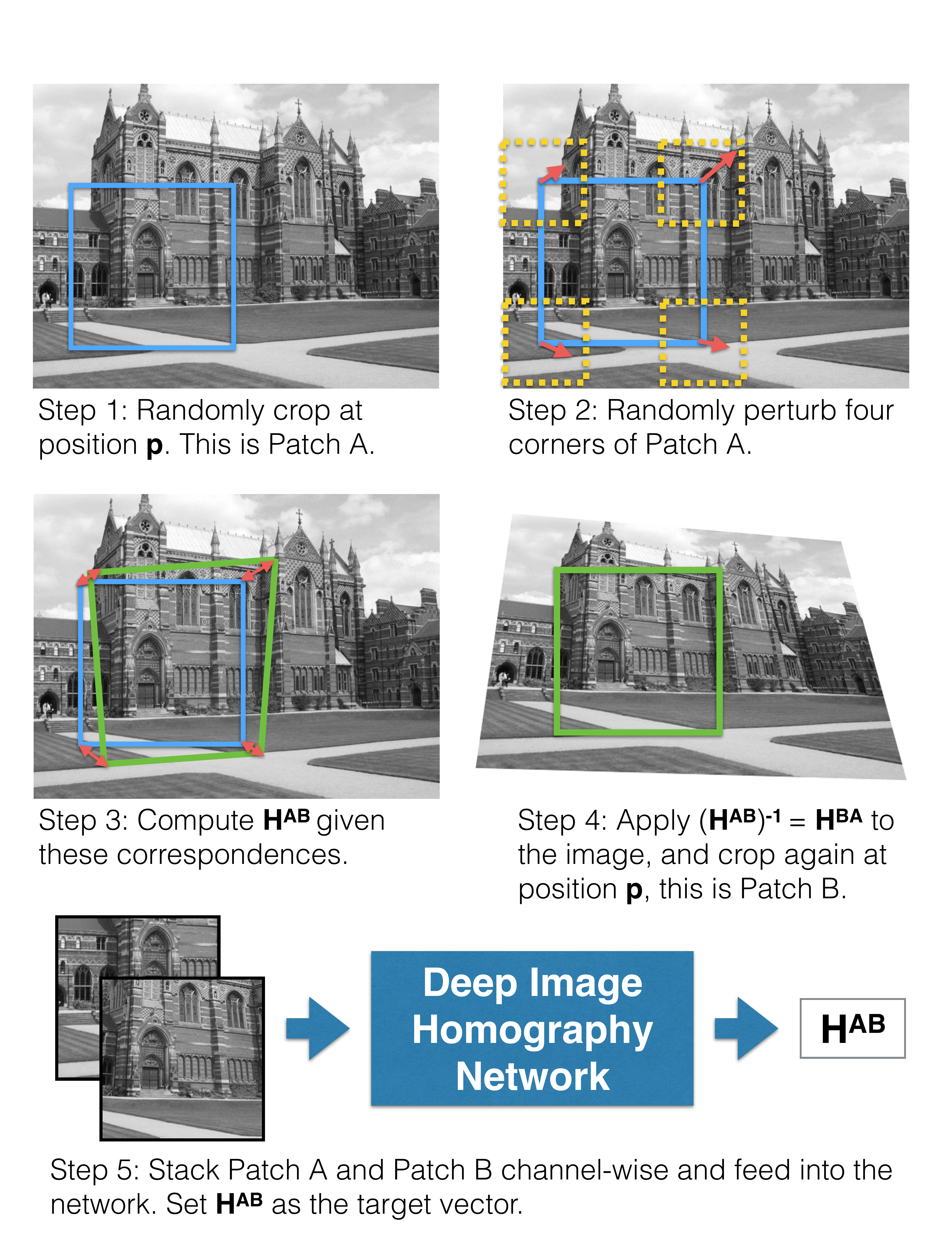}
\caption{{\bf Training Data Generation.} The process for creating
a single training example is detailed. See Section~\ref{sec:data} for more information. \label{fig:data}}
\end{figure}

To generate a single training example, we first randomly crop a square
patch from the larger image $I$ at position $p$ (we avoid the borders to prevent
bordering artifacts later in the data generation pipeline). This random crop is
$I_p$. Then, the four corners of Patch A are randomly perturbed
by values within the range [-$\rho$, $\rho$]. The four correspondences
define a homography  $H^{AB}$. Then, the inverse of this homography
$H^{BA} = (H^{AB})^{-1}$ is applied to the large image to produce image $I'$. A
second patch $I'_p$ is cropped from $I'$ at position $p$. The two grayscale patches, $I_p$ and
$I'_p$ are then stacked channel-wise to create the 2-channel image which is fed directly into our ConvNet. The 4-point parameterization
of $H^{AB}$ is then used as the associated ground-truth training label.



Managing the training image generation pipeline gives
us full control over the kinds of visual effects we want to model. For
example, to make our method more robust to motion blur, we can apply
such blurs to the image in our training set. If we want the method to
be robust to occlusions, we can insert random “occluding” shapes into
our training images. We experimented with in-painting random occluding
rectangles into our training images, as a simple mechanism to simulate real occlusions.

\section{ConvNet Models}
\label{sec:models}

Our networks use 3x3 convolutional blocks with
BatchNorm~\cite{Ioffe15} and ReLUs, and are architecturally similar to
Oxford’s VGG Net~\cite{Simonyan14} (see Figure~\ref{fig:teaser}).
Both networks take as input a two-channel grayscale image sized 128x128x2. In other words,
the two input images, which are related by a homography, are stacked
channel-wise and fed into the network. We use 8 convolutional layers with
a max pooling layer (2x2, stride 2) after every two convolutions. The 8 convolutional
layers have the following number of filters per layer: 64, 64, 64, 64, 128, 128, 128, 128.
The convolutional layers are followed by two fully connected layers. The first
fully connected layer has 1024 units. Dropout with a probability of 0.5 is applied after the
final convolutional layer and the first fully-connected layer. Our two networks
share the same architecture up to the last layer, where the first network produces
real-valued outputs and the second network produces discrete
quantities (see Figure~\ref{fig:convnet}).

The {\bf regression network} directly produces 8 real-valued numbers and
uses the Euclidean (L2) loss as the final layer during training. The advantage of this
formulation is the simplicity; however, without producing any kind of
confidence value for the prediction, such a direct approach could be prohibitive in certain applications.

The {\bf classification network} uses a quantization scheme, has a softmax
at the last layer, and we use the cross entropy loss function during
training. While quantization means that there is some inherent
quantization error, the network is able to produce a confidence for
each of the corners produced by the method. We chose to use 21 quantization
bins for each of the 8 output dimensions, which results in a final layer with
168 output neurons. Figure \ref{confidences} is a visualization
of the corner confidences produced by our method --- notice how the
confidence is not equal for all corners.

\begin{figure*}[ht]
\centering
\includegraphics[width=\textwidth]{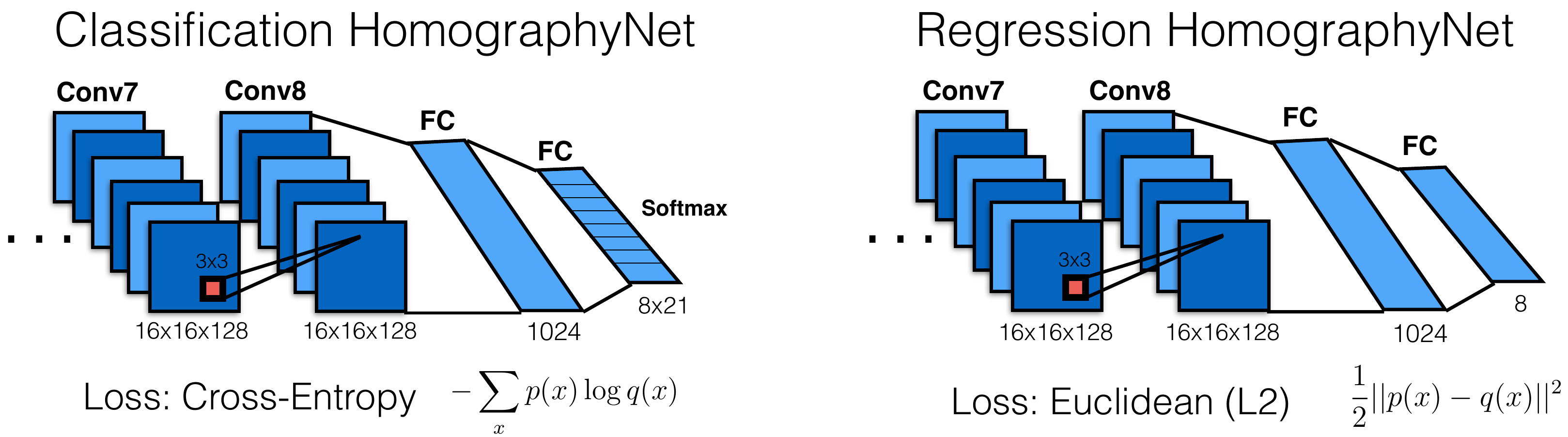}
\caption{{\bf Classification HomographyNet vs Regression HomographyNet.} Our VGG-like Network has 8 convolutional layers and two fully connected layers. The final layer is 8x21 for the classification network and 8x1 for the regression network. The 8x21 output can be interpreted as four 21x21 corner distributions. See Section~\ref{sec:models} for full ConvNet details. \label{fig:convnet}}
\end{figure*}

\section{Experiments}

We train both of our networks for about 8 hours on a single Titan X GPU,
using stochastic gradient descent (SGD) with momentum of 0.9. We use a base
learning rate of 0.005 and decrease the learning rate by a factor of 10 after every
30,000 iterations. The networks are trained for for 90,000 total iterations
using a batch size of 64. We use Caffe~\cite{jia2014caffe}, a popular open-source
deep learning package, for all experiments. 

To create the training data, we use the MS-COCO Training Set. All images are resized to 320x240 and converted to grayscale. We
then generate 500,000 pairs of image patches sized 128x128 related by a homography 
using the method described in Section \ref{sec:data}. We choose $\rho$ = 32,
which means that each corner of the
128x128 grayscale image can be perturbed by a maximum of one quarter of the
total image edge size. We avoid larger random perturbations to avoid extreme transformations.
We did not use any form of pre-training; the weights of the networks were initialized to random values
and trained from scratch. We use the MS-COCO validation set to monitor overfitting, of which
we found very little.

To our knowledge there are no large, publicly available homography estimation test sets, thus
we evaluate our homography estimation approach on our own Warped MS-COCO 14 
Test Set. To create this test set, we randomly chose 5000 images from the test set
and resized each image to grayscale 640x480,
and generate a pairs of image patches sized 256x256 \footnote{We found that very few ORB features were
detected when the patches were sized 128x128, while the HomographyNets had no issues working
at the smaller scale.} and corresponding ground truth homography, using the approach
described in Figure \ref{fig:data} with $\rho$ = 64.

We compare the Classification and Regression variants of the HomographyNet with two
baselines. The first baseline is a classical ORB~\cite{Rublee11} descriptor +
RANSAC + {\verb getPerspectiveTransform() } OpenCV Homography computation.
We use the default OpenCV parameters in the traditional homography estimator. This estimates
ORB features at multiple scales and uses the top 25 scoring matches as input to the RANSAC
estimator. In scenarios where too few ORB features are computed, the ORB+RANSAC approach outputs
an identity estimate. In scenarios where the ORB+RANSAC's estimate is too extreme, the 4-point
homography estimate is clipped at [-64,64]. The second baseline uses a 3x3 identity matrix for every pair of images in the test set.

Since the HomographyNets expect a fixed sized 128x128x2 input, the image pairs from the Warped
MS-COCO 14 Test Set are resized from 256x256x2 to 128x128x2 before being passed through the
network. The 4-point parameterized homography output by the network is then multiplied
by a factor of two to account for this. When evaluating the Classification HomographyNet,
the corner displacement with the highest confidence is chosen.

The results are reported in Figure \ref{fig:eval}. We report the Mean Average Corner Error
for each approach. To measure this metric, one
first computes the L2 distance between the ground truth corner position
and the estimated corner position. The error is averaged over the four corners of the image, and the
mean is computed over the entire test set. While the regression network performs the best, the classification network can produce
confidences and thus a meaningful way to visually debug the results. In certain applications,
it may be critical to have this measure of certainty.



We visualize homography estimations in Figure~\ref{samples}. The blue squares in column 1 are mapped to a
blue quadrilateral in column 2 by a random homography generated from the process described in
Section \ref{sec:data}. The green quadrilateral is the estimated homography. The
more closely the blue and green quadrilateral align, the better. The
red lines show the top scoring matches of ORB features across the
image patches. A similar visualization is shown in columns 3 and 4, 
except the Deep Homography Estimator is used.

\section{Applications}

\begin{figure}[b!]
\centering
\includegraphics[width=.45\textwidth]{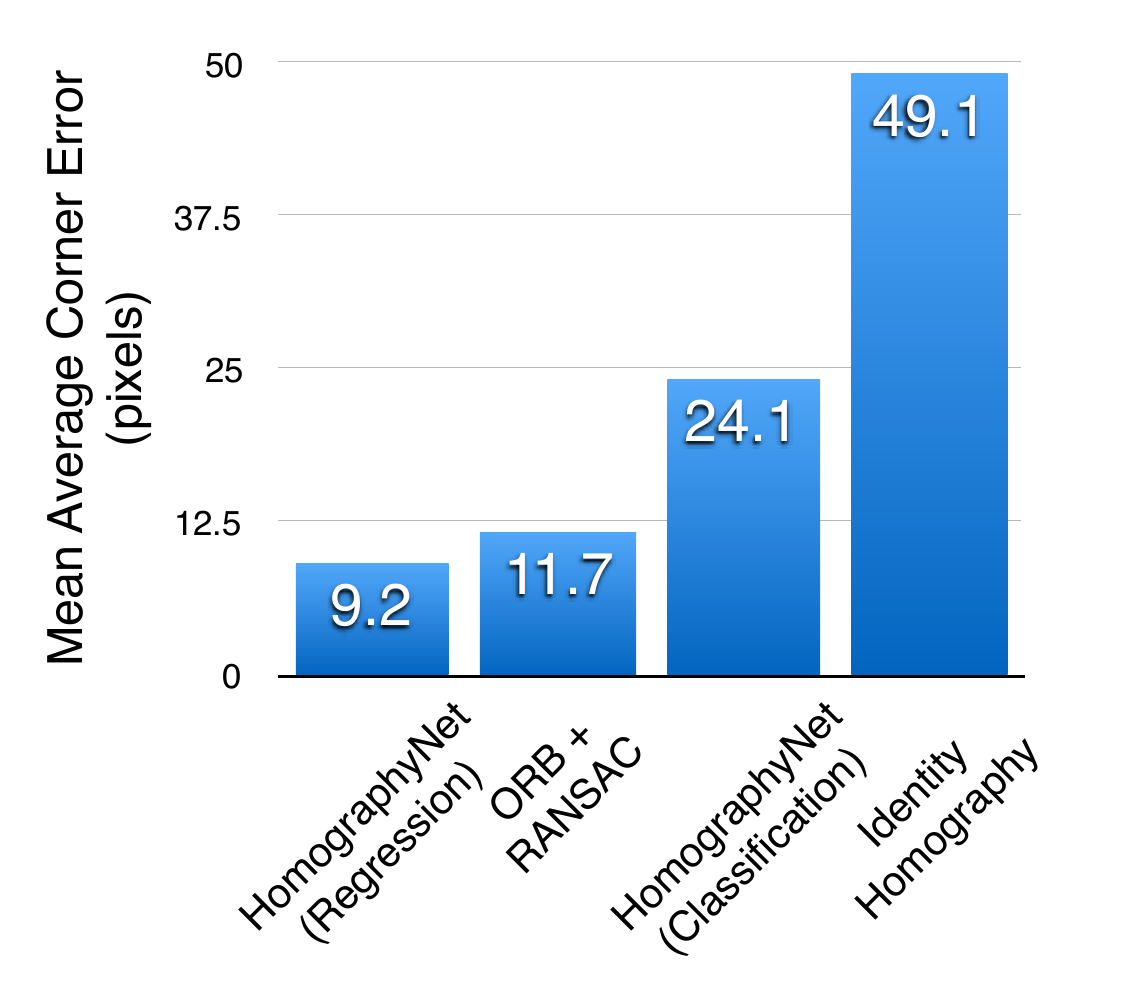}
\caption{ {\bf Homography Estimation Comparison on Warped MS-COCO 14 Test Set.} The mean average corner error is computed for various approaches on the Warped MS-COCO 14 Test Set. The HomographyNet with the regression head performs the best. The far right bar shows the error computed if the identity transformation is estimated for each test pair.} \label{fig:eval}
\end{figure}

Our Deep Homography Estimation system enables a variety of interesting applications. Firstly, our system is fast. It runs at over 300fps with a batch
size of one (i.e. real-time inference mode) on an
NVIDIA Titan X GPU, which enables a host of applications that are simply
not possible with a slower system. The recent emergence of specialized
embedded hardware for deep networks
will enable applications on many embedded systems or 
platforms with limited computational power which cannot afford an expensive and
power-hungry desktop GPU. These embedded systems are capable of running much
larger networks such as AlexNet \cite{Krizhevsky12} in real-time, and should
have no problem running the relatively light-weight HomographyNets.

\begin{figure*}[ht]
\centering
\includegraphics[width=\textwidth]{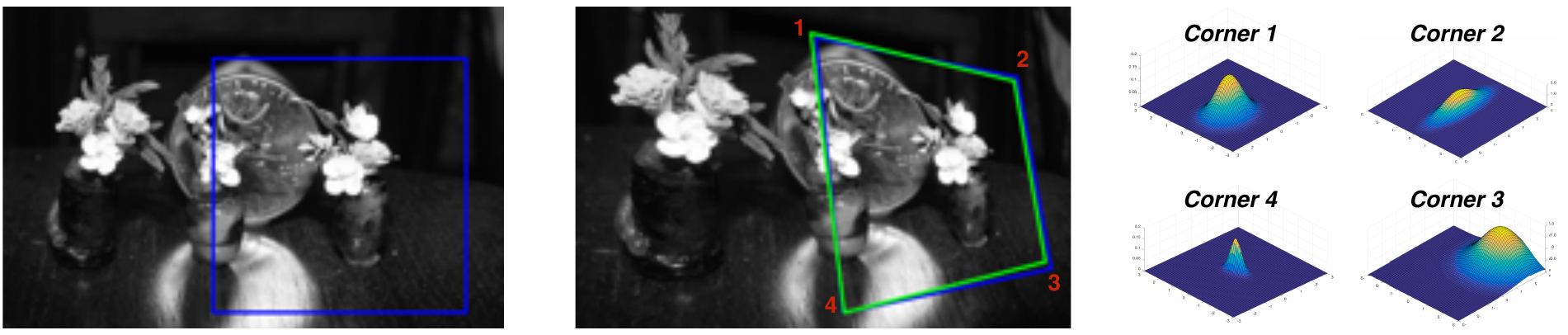}
\caption{{\bf Corner Confidences Measure.} Our Classification HomographyNet produces a score for each potential 2D displacement of each corner. Each corner's 2D grid of scores can be interpreted as a distribution. \label{confidences}} 
\end{figure*}

Secondly, by formulating homography estimation as a machine learning problem, one
can build application-specific homography estimation engines. For example,
a robot that navigates an indoor factory floor using planar SLAM via
homography estimation could be trained solely with images captured from the robot's
image sensor of the indoor factory. While it is possible to optimize a
feature detector such as ORB to work in specific environments, it is not
straightforward. Environment and sensor-specific noise, motion
blur, and occlusions which might restrict the ability of a homography estimation
algorithm can be tackled in a similar fashion using a ConvNet. Other classical
computer vision tasks such as image mosaicing (as in \cite{Szeliski96}) and markerless
camera tracking systems for augmented reality (as in \cite{Simon00}) could also benefit
from HomographyNets trained on image pair examples created from the
target system's sensors and environment.


\section{Conclusion}
\label{sec:conclusion}

In this paper we asked if one of the most essential computer vision
estimation tasks, namely homography estimation, could be cast as a
learning problem. We presented two Convolutional Neural Network
architectures that are able to perform well on this task. Our
end-to-end training pipeline contains two additional insights: using a
4-point “corner parameterization” of homographies, which makes the
parameterization’s coordinates operate on the same scale, and using a
large dataset of real image to synthetically create an seemingly
unlimited-sized training set for homography estimation. We hope that more geometric problems in vision will be tackled using learning paradigms.


\begin{figure*}[ht]
\centering
\includegraphics[width=.99\textwidth]{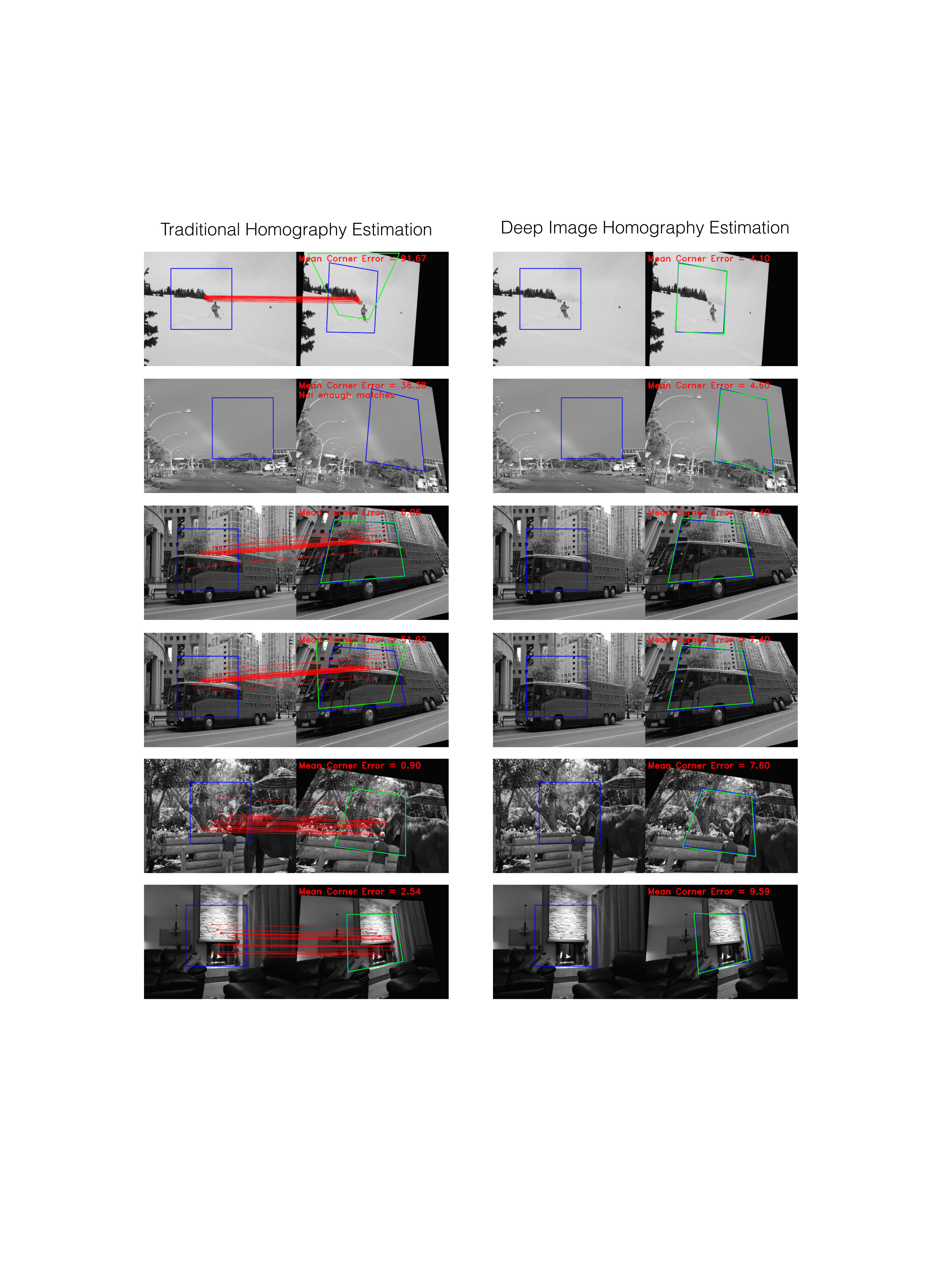}
\caption{ {\small {\bf Traditional Homography Estimation vs
  Deep Image Homography Estimation.} In each of the 12 examples, blue depicts the ground truth region. The left column shows the output of ORB-based Homography Estimation, the matched features in red, and the resulting mapping in green of the cropping. The right column shows the output of the HomographyNet (regression head) in green. Rows 1-2: The ORB features either concentrate on small regions or cannot detect enough features and perform poorly relative to the HomographyNet, which is uneffected by these phenomena. Row 3: Both methods give reasonably good homography estimates. Row 4: A small amount of Gaussian noise is added to the image pair in row 3, deteriorating the results produced by the traditional method, while our method is unaffected by the distortions. Rows 5-6: The traditional approach extracts well-distributed ORB features, and also outperforms the deep method.}
  \label{samples}}
\end{figure*}


\bibliographystyle{plainnat}
\bibliography{references}

\end{document}